\documentclass{fundam}
\usepackage{booktabs}
\usepackage{xcolor}
\usepackage{url}
\usepackage{amsmath}
\usepackage{amsfonts}
\usepackage{graphicx}
\usepackage{todonotes}
\usepackage{wrapfig}
\usepackage{picinpar}
\usepackage[ruled,lined]{algorithm2e}

\begin{document}
\setcounter{page}{1001}
\issue{(2018)}
\title{Classifying and Visualizing Emotions with Emotional DAN}

\author{Ivona Tautkut\.e\\
Polish-Japanese Academy of Information Technology \\
Tooploox \\
Warsaw, Poland\\
s16352@pjwstk.edu.pl
\and Tomasz Trzci\'{n}ski\\
Warsaw University of Technology \\
Tooploox \\
Warsaw, Poland\\
t.trzcinski@ii.pw.edu.pl } \maketitle

\runninghead{I. Tautkut\.e, T. Trzci\'{n}ski}{Classifying and Visualizing Emotions with Emotional DAN}




\begin{abstract}
Classification of human emotions remains an important and challenging task for many computer vision algorithms, especially in the era of humanoid robots which coexist with humans in their everyday life. Currently proposed methods for emotion recognition solve this task using multi-layered convolutional networks that do not explicitly infer any facial features in the classification phase. In this work, we postulate a fundamentally different approach to solve emotion recognition task that relies on incorporating facial landmarks as a part of the classification loss function. To that end, we extend a recently proposed Deep Alignment Network (DAN) with a term related to facial features. Thanks to this simple modification, our model called EmotionalDAN is able to outperform state-of-the-art emotion classification methods on two challenging benchmark dataset by up to 5\%. Furthermore, we visualize image regions analyzed by the network when making a decision and the results indicate that our EmotionalDAN model is able to correctly identify facial landmarks responsible for expressing the emotions.
\end{abstract}


\begin{figure}[ht]
\begin{center}
\includegraphics[width=0.9\textwidth]{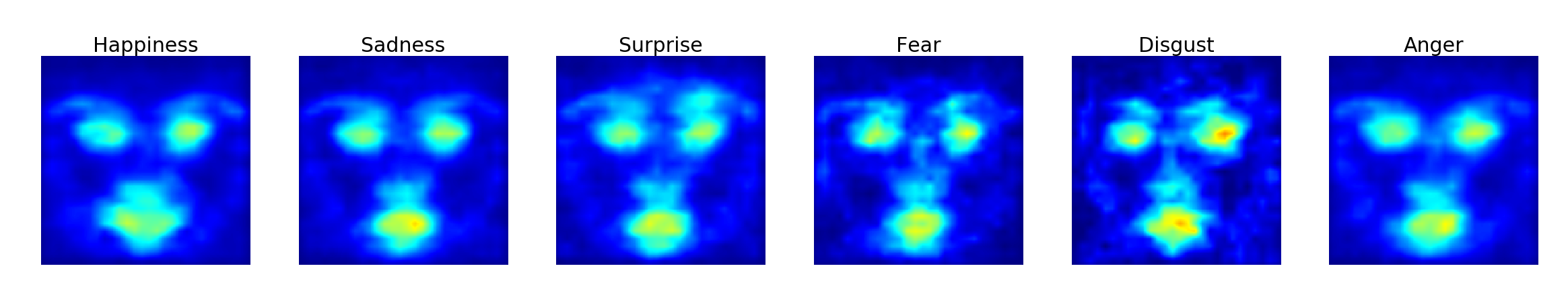}
\end{center}
   \caption{Mean visualisations of Conv 4a layer activations in EmotionalDAN network for emotion classification. Even though during training model has only information about emotion label, spatial information about regions containing information relevant to expressed emotions are correctly captured. Face regions around eyes and mouth influence model's decision the most for all emotion classes. Eyebrow location and strength of activations near nose (e.g Disgust) also show some discriminative information. 
}
\label{fig:header}
\end{figure}

\section{Introduction}

Since autonomous AI systems, such as anthropomorphic robots, start to rapidly enter our lives, their ability to understand social and emotional context of many everyday situations becomes increasingly important. 
One key element that allows the machines to infer this context is their ability to correctly identify human emotions, such as happiness or sorrow. This is a highly challenging task, as people express their emotions in a multitude of ways, depending on their personal characteristics, {\it e.g.} people with an introvert character tend to be more secretive about their emotions, while extroverts show them more openly. Although some simplifications can be applied, for instance reducing the space of recognized emotions or directly applying Facial Action Coding System (FACS) \cite{facs}, there is an intrinsic difficulty embedded in the problem of human emotion classification.

While many Facial Expression Recognition (FER) systems already exist \cite{baseline_inception,baseline_cnn_inception,emotionet,emotionnet2,goingdeeper,3dfer,rnn-fer}, the problem is far from being solved, in particular for expressions that are easily confused when judged without context (e.g \textit{fear} and \textit{surprise}).
Considering facial expression recognition and face alignment are coherently related to each other, they should be beneficial for each other if putting them in a joint framework, e.g facial expression recognition has served as an auxiliary task to enhance landmark localication \cite{sina}.
However, in literature it is rare to see such joint study of the two tasks.
We therefore propose to use a state-of-the-art facial landmark detection model -- Deep Alignment Network (DAN)~\cite{dan} -- and extend it by adding a surrogate term that aims to correctly classify emotions to the neural network loss function. This simple modification allows our method, dubbed EmotionalDAN, to exploit the location of facial landmarks and incorporate this information into the classification process. By training both terms jointly, we obtain state-of-the-art results on two challenging datasets for facial emotion recognition: CK+~\cite{ck} and ISED~\cite{ised}. 

Furthermore, we perform an additional study that aims to interpret decisions made by model. We visualize with gradient-weighted class activation mapping (Grad-CAM \cite{gradcam}) image regions responsible for predicting the concept. Results show that our proposed architecture correctly focuses at the most important for emotion classification regions of the face.

The remainder of this work is organized in the following manner. In Sec.~\ref{sec:related} we discuss related work in facial expression recognition.
In Sec.~\ref{sec:emotionalDAN} we present our approach and introduce EmotionalDAN model. In Sec.~\ref{sec:experiments} we present the datasets used for evaluation, explain in detail how our experiments are performed and present the results compared against baselines. Sec.~\ref{sec:explanations} presents a set of experiments explaining visually the behaviour of our EmotionalDAN model with respect to the facial landmarks. Sec.~\ref{sec:application} illustrates real life application of our proposed model. Finally, Sec.~\ref{sec:conclusions} concludes the paper.

\section{Related work}
\label{sec:related}

It is a common standard to taxonomize human facial movements with Facial Action Coding System published by Ekman and Friesen \cite{facs} in 1978 that describes facial expressions by action units (AUs) based on the anatomy of human face. Out of 30 AUs describing independent movements of the face muscles 12 are related to muscle contractions of upper face and 18 of the lower face. FACS system describes all visible facial muscle movements, and not just those presumed to be related to emotion or any other human state. Although FACS system has been widely used by behavioral scientists and allows for explicit definition of facial expression, it is a tedious task to code by hand.

There has been some work on automatic action units of facial movements system detection. Some methods work on local face patches and explore co-ocurrence of AUs in multilabel setting  \cite{zhao2015} or combine local models by training a different classifier per face region \cite{jaiswal2015}. More recently, attention module has been used for AU detection in weakly superwised manner \cite{shao, shao2018}. Recognizing action units can directly help analyze facial expression \cite{tian2001}.
However even though predicting emotion by detecting presence of AUs provides full transparency of the model, such methods are strictly limited to AU definitions. It has been observed that humans tend to express emotions in a wide variety of facial muscles independent of AUs.

\color{black}
Most of the recently proposed methods for automatic facial expression recognition are Deep Learning based methods and have proven to be more successful at emotion prediction than handcrafted features \cite{CNN_5,emotionet,rnn-fer}. They commonly use some variation of a deep neural network with convolutional layers.
With their broad spectrum of applications to various computer vision tasks, convolutional neural networks (CNN) have also been successful at recognizing emotions. For instance~\cite{CNN_5} propose to use a standard architecture of a CNN with two convolutional, two subsamping and one fully connected layer. Before being processed, the image is spatially normalized with a pre-processing step. Their model achieves state-of-the-art accuracy on CK+~\cite{ck} database of 97.81\%. Some modified versions of this approach also include different numbers of layers (e.g five convolutional layers).

A number of methods is inspired by Inception model~\cite{inception} that achieves state-of-the-art object classification results on the ImageNet dataset~\cite{imagenet}. Inception layers provide an approximation of sparse networks hence are often applied to emotion recognition problem \cite{3dfer}. Ranging from simple transfer learning approaches where Inception-V3 model pretrained on ImageNet~\cite{imagenet} is used with custom softmax classification layer \cite{baseline_inception} to custom architectures with Inception layers \cite{goingdeeper}. In another example \cite{baseline_cnn_inception} propose a deep neural network architecture consisting of two convolutional layers each followed by max pooling and then four Inception layers.

Another method called EmotionNet \cite{emotionet} and its extension EmotionNet2~\cite{emotionnet2} builds up on the ultra-deep ResNet architecture~\cite{DBLP:journals/corr/HeZRS15} and improves the accuracy by using face detection algorithm that reduces the variance caused by a background noise.

Although all the above methods rely on the state-of-the-art deep learning architectures, they draw their inspiration mostly from the analogical models that are successfully used for object classification tasks. We believe that as a result these approaches do not exploit intrinsic characteristics of how humans express emotions, {\it i.e.} by modifying their face expression through moving the landmark features of their faces.  Moreover, vast majority of published methods is evaluated within the same database that the model was trained for with no cross-database comparison. While such accuracy results might be impressive they often lack the ability to generalize to different shooting conditions (lightning, angles, image quality) or subjects of different ethnic backgrounds.


To combine best of the two worlds, we propose a method that is not restricted by limitations of FACS system but by building up on facial landmarks concept can provide insights on how the decisions are made.   

\begin{wrapfigure}{R}{0.4\textwidth}
\centering
\includegraphics[width=0.3\textwidth]{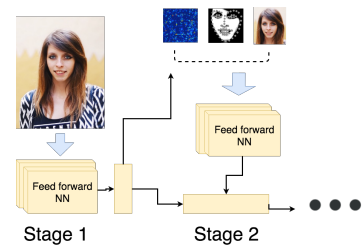}
   \caption{Information about landmark location estimates, landmark heatmaps and mean image are transferred across stages. Landmark location estimates are refined in the proceeding stage.}
\label{fig:architecture-2}
\end{wrapfigure}

\section{EmotionalDAN}
\label{sec:emotionalDAN}
Our approach builds up on the Deep Alignment Network architecture~\cite{dan}, initially proposed  for robust face alignment. The main advantage of DAN over the competing face alignment methods comes from an iterative process of adjusting the locations of facial landmarks. The iterations are incorporated into the neural network architecture, as the information about the landmark locations detected in the previous stage (layer) are transferred to the next stages through the use of facial landmark heatmaps. As a result and contrary to the competing methods, DAN can therefore handle entire face images instead of patches which leads to a significant reduction in head pose variance and improves its performance on a landmark recognition task. DAN ranked $3^{rd}$ in a recent face landmark recognition challenge Menpo~\cite{menpo}.

Originally, DAN was inspired by the Cascade Shape Regression framework and similarly it starts with initial estimate of face shape which is refined after following iterations. In DAN, each iteration is represented with a single stage of deep neural network. During each stage (iteration) features are extracted from entire image instead of local images patches (in contrast to CSR).



Training is composed of consecutive stages where single stage consists of feed-forward neural network and connection layers generating input for next stage. Each stage takes three types of inputs: input image aligned with the canonical shape, features image generated from dense layer of the previous stage and landmarks heatmap.  Therefore output at each DAN stage is defined as:
\begin{equation}
S_t = T^{-1}_t(S_{t-1}) + \Delta S_t,
\end{equation}
where $\Delta S_t$ is the landmarks output at stage $t$ and $T_t$ is the transform that is used to warp the input image to canonical pose.

In this work, we hypothesize that DAN's ability to handle images with large variation and provide robust information about facial landmarks transfers well to the task of emotion recognition.
To that end, we extend the network learning task with an additional goal of estimating expressed facial emotions. We incarnate this idea by modifying the loss function with a surrogate term that addresses specifically emotion recognition task and we minimize both landmark location and emotion recognition terms jointly. The resulting loss function $\mathcal{L}$ can be therefore expressed as:

\begin{equation}
\label{loss}
\mathcal{L} = \alpha \cdot \frac{\parallel S_t - S^{*} \parallel }{d} - \beta \cdot E^{*} \cdot log(E_{t} ) ,
\end{equation}
\noindent where $S_{t}$ is the transformed output of predicted facial landmarks at stage $t$,
$E$ is the softmax output for emotion prediction.  $S^{*}$ is the vector of ground truth landmark locations, $d$ is the distance between the pupils of ground truth that serves as a normalization scalar and $E^{*}$ is the ground truth for emotion labels. We weigh the influence of the terms with $\alpha$ and $\beta$ coefficients. 

We present the final version of our network in the table \ref{table:architecture}. It was originally inspired by network used in ImageNet ILSVRC competition (2014) \cite{Simonyan} and contains four convolutional layer pairs followed by pooling layers. Top layers of the network consist of one common fully connected layer and two separate fully connected layers for landmark and emotion features.



\section{Experiments}
\label{sec:experiments}
In this section we perform quantitative evaluation of our model against published baselines as well as present an overview of datasets used for training and testing.

\begin{wrapfigure}{R}{0.4\textwidth}
\centering
\includegraphics[width=0.3\textwidth]{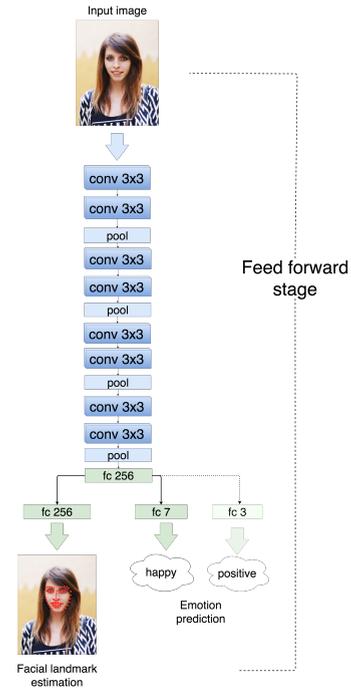}
  \caption{Visualisation of single stage of EmotionalDAN. There are two independent fully connected layers for the task of facial landmarks localization and emotion prediction. Size of the second one depends on number of emotion classes in the model.}
\label{fig:architecture}
\end{wrapfigure}

\subsection{Datasets}

We include datasets that are made available to the public (upon request) and present a high variety of subjects' ethnicity. All compared models are trained on AffectNet \cite{affectnet} and evaluated cross-database on remaining test sets.

\textbf{AffectNet} \cite{affectnet} is by far the largest available database for facial expression. It contains more than 1,000,000 facial images from the Internet collected by querying major search engines with emotion related keywords. About half of the retrieved images were manually annotated for the presence of seven main facial expressions. 7 000 images from AffectNet database are set aside for validation and test sets. 

\textbf{CK+} \cite{ck} includes both posed and non-posed (spontaneous) expressions. 123 subjects are photographed in 6 prototypic emotions. For our analysis we only include images with validated emotion labels.

\textbf{JAFFE} \cite{jaffe} The database contains 213 images of 7 facial expressions (6 basic facial expressions + 1 neutral) posed by 10 Japanese female models. Each image has been rated on 6 emotion adjectives by 60 Japanese subjects.

\textbf{ISED} \cite{ised} Indian Spontaneous Expression Database. Near frontal face video was recorded for 50 participants while watching emotional video clips. The  novel  experiment  design  induced  spontaneous  emotions  among  the  participants  and  simultaneously gathered their self  ratings  of experienced  emotion. For evaluation individual frames from recorded videos are used.



\begin{table}
\setlength{\tabcolsep}{10pt}
\centering
\caption{Structure of the feed-forward part of EmotionalDAN network stage with multiple outputs. Dimensions of the last fully connected layer before emotion classification depend on the number of emotion classes used in training.}
\begin{tabular}{|c | c  | c | c|} 
 \hline
 Name & Input shape & Output shape & Kernel \\ 
 \hline
 conv1a & 224$\times$224$\times$1 & 224$\times$224$\times$64 & 3$\times$3,1,1 \\ 
 conv1b & 224$\times$224$\times$64 & 224$\times$224$\times$64 & 3$\times$3,64,1 \\
 pool1 & 224$\times$224$\times$64 & 112$\times$112$\times$64 & 2$\times$2,1,2 \\
 conv2a & 112$\times$112$\times$64 & 112$\times$112$\times$128 & 3$\times$3,64,1 \\
 conv2b & 112$\times$112$\times$128 & 112$\times$112$\times$128 & 3$\times$3,128,1 \\  
 pool2 & 112$\times$112$\times$128 & 56$\times$56$\times$128 & 2$\times$2,1,2 \\
 conv3a & 56$\times$56$\times$128 & 56$\times$56$\times$256 & 3$\times$3,128,1 \\
 conv3b & 56$\times$56$\times$256 & 56$\times$56$\times$256 & 3$\times$3,256,1 \\
 pool3 & 56$\times$56$\times$256 & 28$\times$28$\times$256 & 2$\times$2,1,2 \\
 conv4a & 28$\times$28$\times$256 & 28$\times$28$\times$512 & 3$\times$3,256,1 \\
 conv4b & 28$\times$28$\times$512 & 28$\times$28$\times$512 & 3$\times$3,512,1 \\
 pool4 & 28$\times$28$\times$512 & 14$\times$14$\times$512 & 2$\times$2,1,2 \\
 fc1 & 14$\times$14$\times$512 & 1$\times$1$\times$256 & - \\
 \hline
 fc2\_landmark & 1$\times$1$\times$256 & 1$\times$1$\times$136 & - \\
 fc2\_emotion & 1$\times$1$\times$256 & 1$\times$1$\times$\{3,7\} & - \\
 
\hline
\end{tabular}
\label{table:architecture}
\end{table}

\subsection{Datasets preparation and training}

To allow for fair comparison we follow an unified approach for all datasets and methods.

While some datasets come with ground-truth information about bounding boxes of present faces (AffectNet), most test sets do not contain such information. Face regions often account only for small part of the image with a lot of unnecessary background (ISED). To address this issue, we extracted regions of interest with face detection algorithm Multi-task CNN \cite{mtcnn}. For $<2\%$ of test images where algorithm failed to recognize a face, full images were used.
All test images were resized to $224 \times 224$, converted to black and white, and normalized by mean and standard deviation of the training set.

Training procedure is independent for three and seven emotion classes. For simplified emotions we perform a mapping of original ground truth labels where \textit{fear, sadness, disgust} and \textit{anger} are mapped to \textit{negative} emotion, \textit{happiness} and \textit{contempt} to \textit{positive} emotion and \textit{neutral} class is kept without change. In this case we do not include images labeled with \textit{surprise} as this emotion might have both positive and negative connotations.

Similarly to \cite{dan}, training of EmotionalDAN is performed sequentially - first stage is trained until validation error stops improving. Afterwards, second stage is added and trained. After an initial set of experiments we set $\alpha$ and $\beta$ coefficients to $\alpha=0.4$ and $\beta=0.6$. To prevent overfitting we add dropout layers with $p=0.5$ for all stages after pooling layers. To improve training procedure, we use cyclical learning rate with triangular policy where learning rate varies between $ 0.0001$ (\textit{base\_lr}) and $0.05$ (\textit{max\_lr}).
Code with EmotionalDAN acrhitecture and training details is publicly available in GitHub repository\footnote{https://github.com/IvonaTau/emotionaldan}.

\color{black}

\begin{table}
\setlength{\tabcolsep}{20pt}
\def\arraystretch{1.1}
\centering
\caption{Cross-database accuracy results compared for different model architectures and seven emotion categories. All models are trained on AffectNet database. Face detection is applied as a preprocessing step on all test sets for all methods.}
\label{results-accuracy-7}
\begin{tabular}{@{}c| cccc@{}}
\toprule
          & CK+ & JAFFE & ISED &  \\ \midrule
CNN (2)  & 0.628  &   0.484   &  0.516  \\
CNN (5) &  0.728  &  \textbf{0.502}  &  0.593 \\
Inception-V3 &  0.304   &   0.268    & 0.479 \\
EmotionNet 2 & 0.204   &   0.249   & 0.21\\
\textbf{EmotionalDAN} & \textbf{0.736}  &  0.465 &  \textbf{0.62} \\
\bottomrule
\end{tabular}
\end{table}

\subsection{Results}
Tables~\ref{results-accuracy-7} and \ref{results-accuracy-3} show the results of the evaluation of our EmotionalDAN method and the competing approaches. Although the accuracy varies between the tested datasets, our approach outperforms the competitors by a large factor of up to 5\% on two out of three benchmark datasets, namely on CK+ and ISED. The performance of our method is inferior to convolutional neural networks on the JAFFE dataset, although the accuracy values obtained on this dataset are generally lower than the competitors. We believe that this may be the result of a more challenging image acquisition conditions. Furthermore, our results show that convolutional neural networks achieve competitive results when compared with other methods despite their simplistic architecture.

\begin{figure}
\begin{center}
\includegraphics[width=0.6\textwidth]{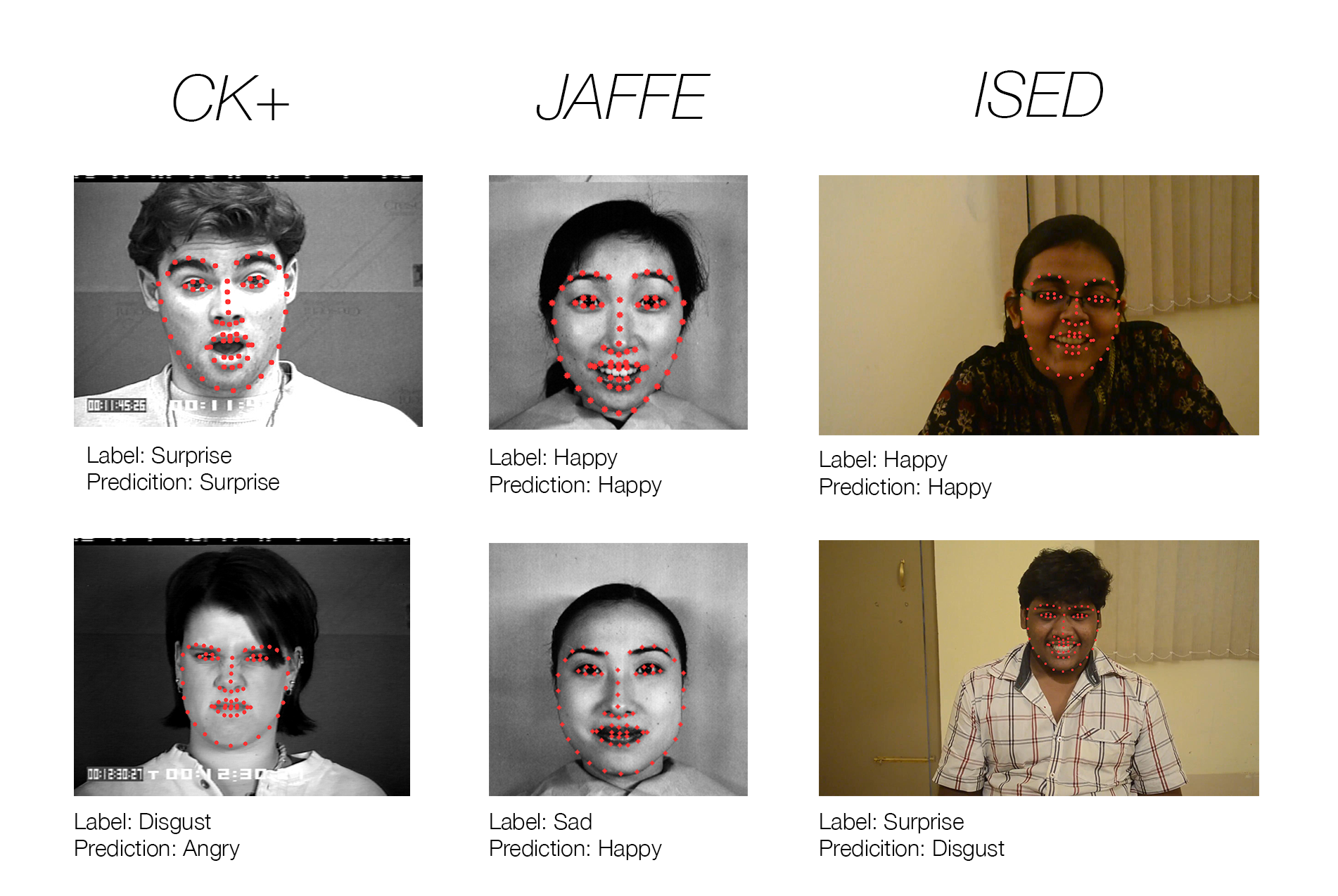}
\end{center}
   \caption{Mapping of EmotionDAN predictions to original images from evaluated test sets. The top row shows examples of correct predictions while the bottom one illustrates classification errors. Most of the errors happen when ambiguous emotions are expre}
\label{fig:emotions}
\end{figure}

\begin{table}
\setlength{\tabcolsep}{20pt}
\def\arraystretch{1.1}
\centering
\caption{Cross-database accuracy results compared for different model architectures and three emotion categories - positive, negative and neutral.}
\label{results-accuracy-3}
\begin{tabular}{@{}c| cccc@{}}
\toprule
          & CK+ & JAFFE & ISED &  \\ \midrule
CNN (2)     &  0.819   &  0.525   &  0.814  \\
CNN (5)  &  0.92  & \textbf{0.765}  & 0.867\\
Inception-V3 &  0.582  &  0.536  & 0.673 \\
EmotionNet 2 & 0.478   &   0.497   & 0.587\\
\textbf{EmotionalDAN} & \textbf{0.921} & 0.634 & \textbf{0.896} \\
\bottomrule
\end{tabular}
\end{table}

Qualitative results are presented in Fig.\ref{fig:emotions} and show examples of correct and incorrect predictions for each testset.

\section{Visual Explanations}
\label{sec:explanations}

In this section we present visual explanations for emotion classification with EmotionalDAN.

\subsection{Grad-CAM}

To gain more insights from our model, we produce visual explanations for classification decisions using a popular gradient-based localization technique Grad-CAM \cite{gradcam}. 
This approach produces a coarse localization map of gradients flowing into the final convolution layer of arbitrary CNN architecture. Neurons in upper convolutional layers look for class-specific semantic information in the image, information that is lost in proceeding fully connected layers. 

Specifically, class-discriminative localization map Grad-CAM $L^C_{Grad-CAM} \in \mathbb{R}^{u \times v}$ of width $u$ and height $v$ for class $c$ is obtained by first computing the gradient of the score for class $c$ with respect to feature maps $A^k$ of convolutional layer. Then, neuron importance weights are obtained by global-average-pooling these gradients flowing back:

\begin{equation}
\alpha^c_k = \frac{1}{Z} \sum_i \sum_j \frac{\partial y^c}{\partial A^k_{i,j}}
\end{equation}

Finally, weighted combination of forward activation maps is followed by ReLU:

\begin{equation}
L^C_{Grad-CAM} = ReLU \left( \sum_k \alpha^c_k A^k \right).
\end{equation}

\begin{figure}
\begin{center}
\includegraphics[width=1.0\textwidth]{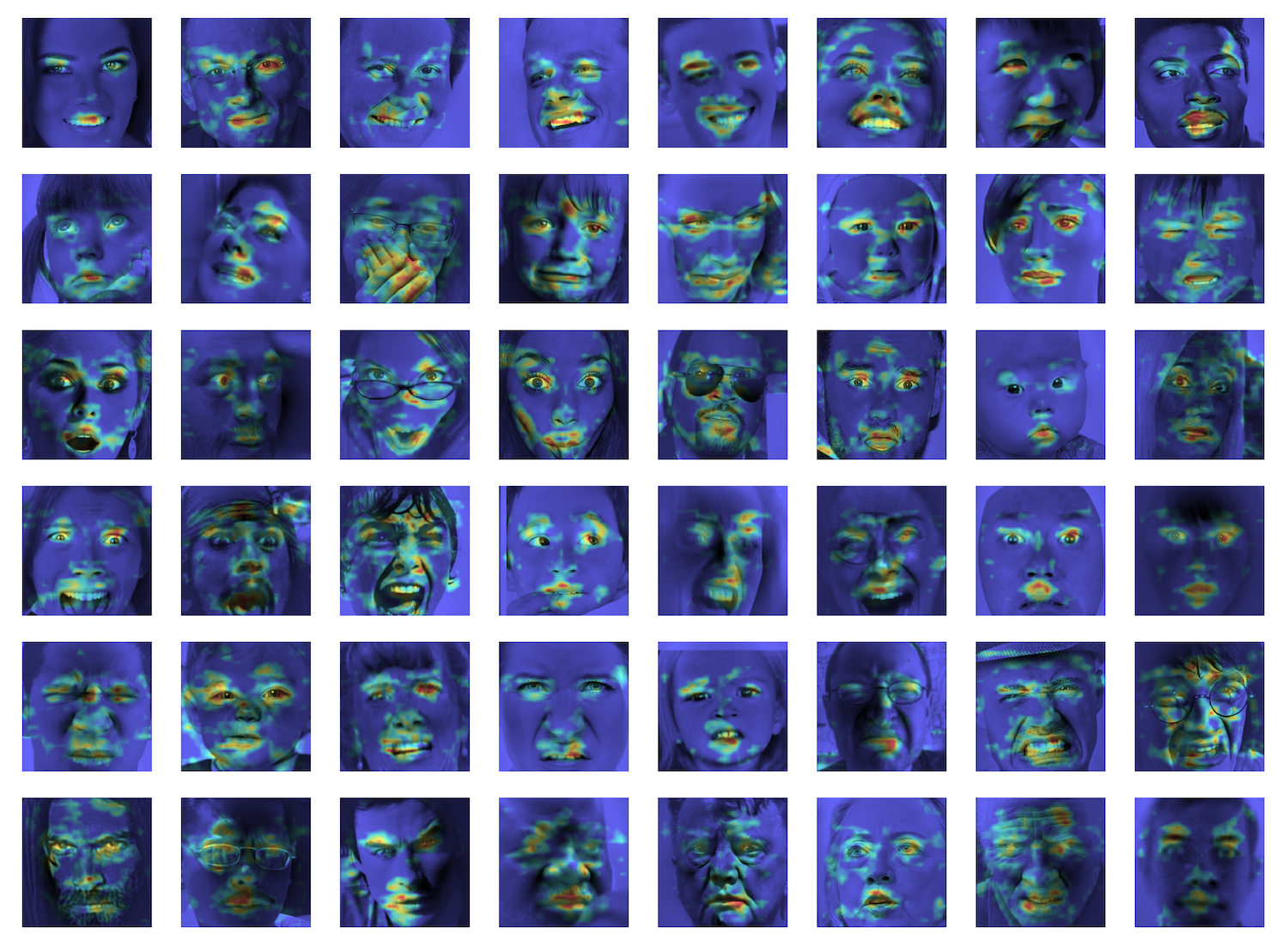}
\end{center}
   \caption{Grad-CAM visual explanations for emotion classification on test set from AffectNet database. Each row represents random sample of images for emotion label (happiness, sadness, surprise, fear, disgust, anger). EmotionalDAN is able to capture important information from face regions close to eyes, brows or mouth.}
\label{fig:gradcams}
\end{figure}

We present a visualization of Grad-CAM activations for final convolutional layer (Conv 4a) of EmotionalDAN in Fig. \ref{fig:gradcams}. Most of the decisions related to emotion classification is judged based on the regions surrounding mouth, eyes, nose and brows. The model correctly identifies those regions although no prior information about what defines a given emotion was known to the network.

\subsection{Grad-CAM activation analysis per emotion label}

We perform a detailed analysis of Grad-CAM activations for different emotion labels. We take a subset of AffectNet test set $I^{test}$ such that we only have images with faces facing forward. More formally, we restrict conditions on eye corner locations by taking the following subset of test images:

\begin{equation}
  I^{front} = \left\{ I \in I^{test} \quad s.t \quad  \lVert (x,y)_{left\ eye} - (x_l,y_l) \rVert < \epsilon \wedge    \lVert  (x,y)_{right\ eye} - (x_r,y_r) \rVert < \epsilon \right\}
\end{equation}

where $(x_l, y_l)$ are mean coordinates of left eye left corner and $(x_r, y_r)$ are mean coordinates of right eye right corner.

For each emotion class $C$:
\begin{equation}
I^C = \left\{ I \in I^{front} \quad s.t \quad y(I) = C \right\}
\end{equation}

\begin{figure}
\begin{center}
\noindent\includegraphics[width=1.0\textwidth]{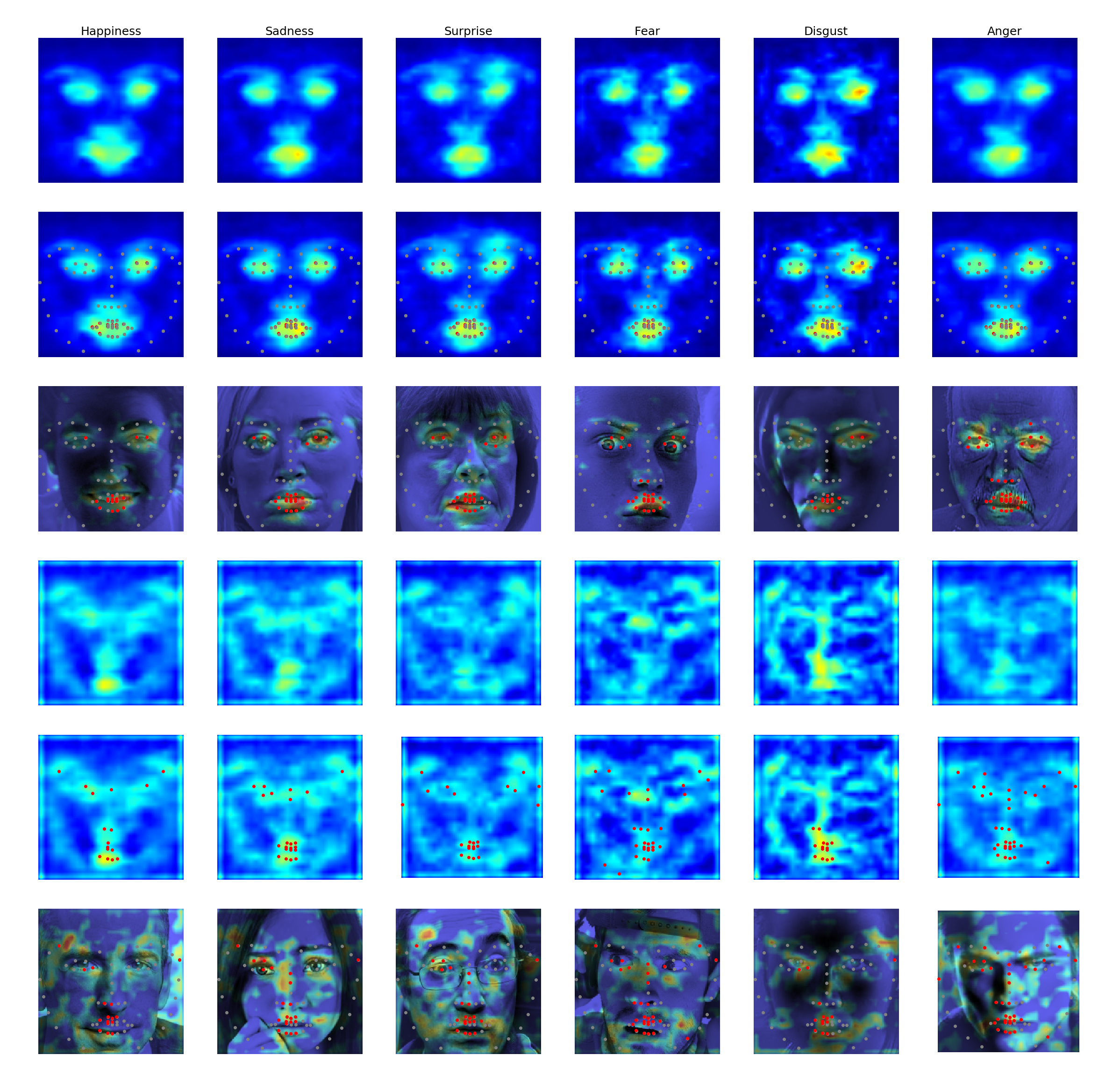}
\end{center}
   \caption{Generated mean Grad-CAM visualizations for each emotion label. First three rows represent heatmaps for Conv 4a layer of EmotionalDAN. The next three were generated with activations from Conv 4b layer. First and fourth rows represent mean Grad-CAM activation heatmaps. Second and fifth present most activated facial landmarks (in red). In third and sixth, images with closest Grad-CAM activations to the mean heatmap are shown.}
\label{fig:mean-gradcams}
\end{figure}

For each set $I^C$ we calculate mean localization map:  $$\overline{L}^C_{Grad-CAM} = \frac{1}{N_C} \sum_{i \in I^C} L^{(i)}_{Grad-CAM}$$

where $N_C$ is the cardinality of $I^C$. Figure \ref{fig:mean-gradcams} shows heatmaps of mean localization maps for two last convolutional layers in EmotionalDAN architecture: Conv4a and Conv4b. Penultimate convolutional layer Conv4a shows a more focused activation of most important face regions - mouth and eyes.

Going further, we extract most activated regions in each mean localization map and compare them to Emotional Facial Action Coding System (EMFACS \cite{emfacs}). To that end, we use AU descriptions to relate them to facial landmarks. Overview is presented in Table \ref{au-landmarks}. Due to transient nature of AUs, the relationship between AU and related facial landmarks is not a strong one. It however indicates points of interest where information about expressed emotion should be located. For example, \textit{Happiness} is documented as presence of AU6 and AU12, where AU6 is \textit{Cheek Raiser} and 
AU12 is \textit{Lip Corner Puller} \cite{emfacs}. Hence, to detect happiness one should focus its attention on face regions close to cheeks and lips.

To verify this approach, for each emotion we retrieve top $k$ activated landmarks, where $k$ is the total number of related landmarks for given emotion from Table \ref{au-landmarks}. Visualization of mean activated landmarks is presented in Figure \ref{fig:mean-gradcams}. We then calculate overlap between most activated landmarks in our model and landmarks from Table \ref{au-landmarks}. We present detailed results in Table \ref{results-mean-gradcams}. Layer Conv 4a gives slightly closer results to AU related landmarks than Conv 4b.

\begin{table}
\setlength{\tabcolsep}{7pt}
\def\arraystretch{1.0}
\centering
\caption{Facial expression descriptions using EMFACS \cite{emfacs} and their relation to facial landmarks.}
\label{au-landmarks}
\begin{tabular}{c|c|c| c}
\toprule
 Emotion & Related AUs & AU description & Related Facial Landmarks \\ \midrule
 
Happiness  & 6 & Cheek Raiser & 1,2 ,14,15  \\
& 12 & Lip Corner Puller & 48, 49, 53, 54, 55, 59, 60, 64 \\

Sadness & 1 & Inner Brow Raiser & 17, 18, 19, 20, 21  \\
& 4 & Brow Lowerer & 22, 23, 24, 25, 26 \\
& 15 & Lip Corner Depressor &  48, 49, 53, 54, 55, 59, 60, 64  \\ 

Surprise & 1 & Inner Brow Raiser &  20, 21, 22, 23 \\
& 2 & Outer Brow Raiser & 17, 18, 19,  24,  25, 26  \\
& 5 & Upper Lid Raiser & 37, 38, 39, 42, 43, 44 \\ 
& 26 & Jaw Drop & 55, 56, 57, 58, 59, 60, 61,  \\
& & & 62, 63, 64, 65, 66, 67\\

Fear & 1 & Inner Brow Raiser &  20, 21, 22, 23\\
&  2 & Outer Brow Raiser &  17, 18, 19,  24,  25, 26 \\ 
& 4 & Brow Lowerer & 17, 18, 19, 20, 21, 22, 23, 24, 25, 26 \\
& 5, 7 & Upper Lid Raiser, Lid Tightener &  37, 38, 39, 42, 43, 44  \\
& 20 & Lip Stretcher & 48, 49, 53, 54, 55, 59, 60, 64  \\
& 26 & Jaw Drop & 55, 56, 57, 58, 59, 60, 61, \\
& & & 62, 63, 64, 65, 66, 67\\

Disgust & 9 & Nose Wrinkler & 27, 28, 29, 30, 31, 32, 33, 34, 35 \\ 
& 15 & Lip Corner Depressor &  48, 49, 53, 54, 55, 59, 60, 64 \\
& 16 & Lower Lip Depressor & 48, 54, 55, 56, 57, 58, 59, 60, 64 \\

Anger & 4 & Brow Lowerer &  17, 18, 19, 20, 21, 22, 23, 24, 25, 26 \\
& 5, 7 & Upper Lid Raiser, Lid Tightener &  37, 38, 39, 42, 43, 44 \\ 
& 23 & Lip Tightener & 48, 49, 50, 51, 52, 53, 54, 55, 56, 57, \\
& & & 58, 59,60, 61, 62, 63, 64, 65, 66, 67\\

\bottomrule
\end{tabular}
\end{table}

\begin{table}
\setlength{\tabcolsep}{8pt}
\def\arraystretch{1.0}
\centering
\caption{Accuracy for top activated landmarks with Grad-CAM when compared to Action Units (AU) related landmarks of given emotion. Outputs of different Emotional DAN final convolutional layers are compared.}
\label{results-mean-gradcams}
\begin{tabular}{@{}c| cccccc | c @{}}
\toprule
 & Happy & Sad & Surprise & Fear & Disgust & Anger & Avg \\ \midrule
Conv 4a    &  0.375 &  0.455  &  0.522 & 0.633 &0.214 & 0.647 & \textbf{0.474} \\
Conv 4b  &  0.312  & 0.409 & 0.478 & 0.6 & 0.429 & 0.559 & 0.464 \\
\bottomrule
\end{tabular}
\end{table}

\color{black}

\vspace{-0.1cm}
\section{Application}
\label{sec:application}
\begin{figure}[t!]
\begin{center}
\includegraphics[width=0.5\textwidth]{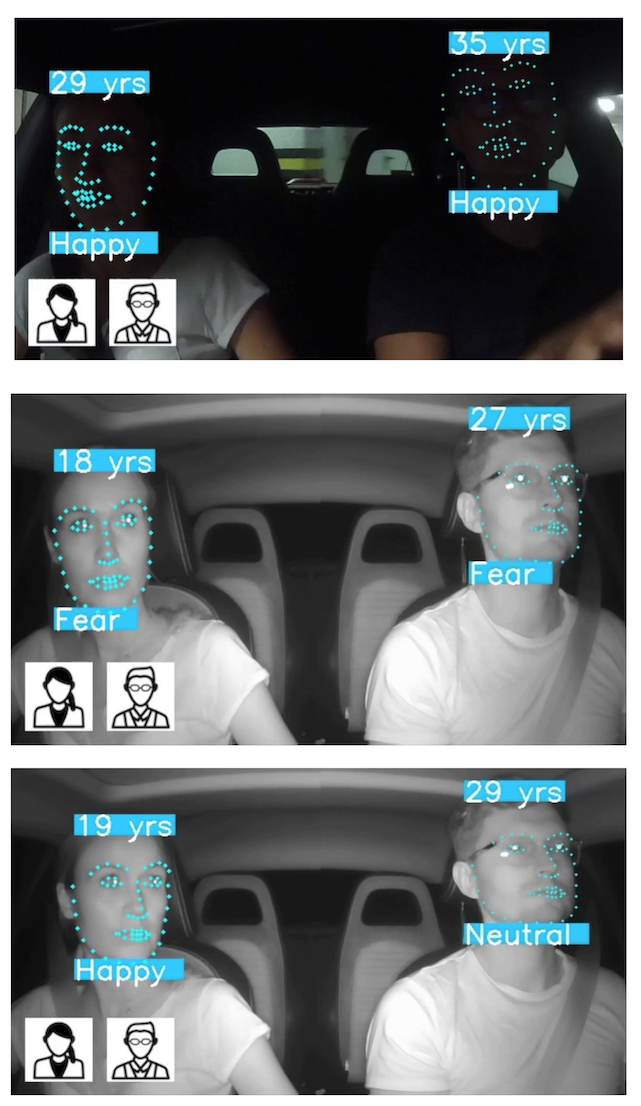}
\end{center}
   \caption{Our emotion recognition model in passenger detection system for autonomous cars. Emotion recognition is performed on detected facial regions.}
\label{fig:car}
\end{figure}

We implement our emotion recognition model as a part of the in-car analytics system to be deployed in autonomous cars. Figure~\ref{fig:car} shows the results obtained by the camera installed inside a car. As autonomous car operation can potentially be influenced by emotions of the passengers ({\it e.g.} fear of speed expressed on passenger's face could signal the need for speed reduction), this is an excellent playground for our method to show its full potential. Although alternative applications are possible, we believe that this use case showcases the capabilities of our method and can serve as an interesting input to the driving system, typically focused on the exterior views from outside the car. 

\section{Conclusion}
\label{sec:conclusions}

In this paper, we overview extension of our previous method \cite{tautkute_cvpr} for emotion recognition that allows to exploit facial landmarks. Although the results computed on the JAFFE dataset show that there is still place for improvement, we believe that this approach has a strong potential to outperform currently proposed methods. In future work, we will therefore focus on improving our method by using attention mechanism on facial landmarks and experiment with additional loss function terms. We also plan to investigate other applications of our method, {\it e.g.} in the context of autistic children with incapabilities related to emotion recognition.

%
\bibliographystyle{ieeetr}
\bibliography{bibliography.bib}

\end{document}